\documentclass[sigconf]{acmart}

\usepackage{cleveref}
\usepackage{tikz}
\usetikzlibrary{positioning, shapes.multipart, fit, backgrounds, calc}

\AtBeginDocument{%
  }

\copyrightyear{2026}
\acmYear{2026}
\setcopyright{cc}
\setcctype{by}
\acmConference[CIKM '26]{Proceedings of the 35th ACM International Conference on Information and Knowledge Management}{November 07--11, 2026}{Rome, Italy}
\acmBooktitle{Proceedings of the 35th ACM International Conference on Information and Knowledge Management (CIKM '26), November 07--11, 2026, Rome, Italy}
\acmDOI{10.1145/3799682.3840879}
\acmISBN{979-8-4007-2539-5/2026/11}

\begin{document}

\title[Anchoring Bias in LLM-as-a-Judge Systems]{Anchoring Bias in LLM-as-a-Judge Systems:\\Prior Scores Compromise Evaluation Independence}

\author{Ante Kapetanovic}
\email{ankapetanovic@infobip.com}
\affiliation{%
  \institution{Infobip}
  \city{Split}
  \country{Croatia}
}

\author{Kemal Altwlkany}
\email{kaltwlkany@infobip.com}
\affiliation{%
  \institution{Infobip}
  \city{Sarajevo}
  \country{Bosnia and Herzegovina}
}

\author{Andro Mercep}
\email{amercep@infobip.com}
\affiliation{%
  \institution{Infobip}
  \city{Zagreb}
  \country{Croatia}
}

\author{Tomislav Duricic}
\email{tduricic@infobip.com}
\affiliation{%
  \institution{Infobip}
  \city{Zagreb}
  \country{Croatia}
}

\author{Emanuel Lacic}
\email{emlacic@infobip.com}
\affiliation{%
  \institution{Infobip}
  \city{Zagreb}
  \country{Croatia}
}

\renewcommand{\shortauthors}{Ante Kapetanovic, Kemal Altwlkany, Andro Mercep, Tomislav Duricic, and Emanuel Lacic}


\begin{abstract}
  Large language models (LLMs) increasingly assess generated content, giving rise to the LLM-as-a-Judge paradigm.
  These systems now score outputs, filter content, and gate iterative refinement in production pipelines, where each judgment is often assumed to be independent of earlier evaluations.
  We test this assumption using three prompt conditions: no metadata, revision framing, and anchored metadata containing revision, attempt, and prior-score fields.
  We show that prior scores, even when included only as context metadata, anchor judgments and systematically shift ratings toward their values.
  Across 192{,}000 attempted evaluations (185{,}271 successful) seven out of the eight evaluated models have 95\% task-stratified bootstrap intervals below zero for the total anchored-metadata effect on 20 fixed texts.
  Cohen's $d$, a standardized measure of the difference between score distributions, reaches an absolute value of 0.71.
  Token-level analysis of selected model-task probes suggests a threshold-like response pattern: introducing anchored metadata produces a marked redistribution of output-score probabilities, while changing the anchor value within the tested below-threshold range produces comparatively little additional variation.
  On categorical industry data with human-labeled ground truth, anchored metadata blocks 48\% of error corrections and flips 10.18\% of correct judgments toward assigned wrong label, demonstrating the bias extends beyond numerical scoring to categorical decisions.
  Neither Chain-of-Thought nor a metadata-disregard warning reduces the total effect, although warning improves the paired accuracy effect relative to baseline in the industry experiment.
  Reliable LLM evaluation demands careful context engineering rather than passive assumption of impartiality, where effective mitigation must be validated for intended model, and specific task or domain.
\end{abstract}


\settopmatter{printacmref=true}

\maketitle

\section{Introduction}
\label{sec:introduction}

Large language models (LLMs) can serve as automated evaluators of generated content, replacing or supplementing human judgment in what is known as the LLM-as-a-Judge paradigm~\citep{gu2026survey}.
Standardized benchmarks and frameworks such as MT-Bench~\citep{zheng2023judging} and G-Eval~\citep{liu2023geval} adopt this approach, treating the judge as an independent scorer.
In production, LLM judges score model outputs for preference learning and content moderation, gate iterative refinement loops~\citep{madaan2023selfrefine,bao2025iterative}, and arbitrate between agents in multi-step workflows~\citep{chhetri2025structsense,yuksel2025multiagent}.

In each of these settings, a judge's prior score can persist in the context window of a subsequent call, whether as a metadata tag on a revision, as an upstream label consumed by a downstream stage, or as the result of a previous round of self-evaluation.
The implicit assumption is that each judgment remains independent of those that came before.

A natural starting point is the cognitive-bias literature on humans.
\citet{tversky1974judgment} described \emph{anchoring}, the tendency for numerical estimates to be pulled toward an initially observed value even when that value is uninformative.
In their canonical demonstration, participants estimated the percentage of African countries in the UN at 25\% versus 45\% depending on whether a spun wheel-of-fortune showed 10 or 65 just before the question.
The proposed mechanism is anchoring-and-adjustment, in which estimators take the visible value as a starting point and adjust insufficiently away from it~\citep{lieder2018anchoring}.
\citet{tversky1981framing} later demonstrated \emph{framing} effects, where equivalent options described in terms of gains rather than losses produce systematically different choices. Participants overwhelmingly preferred a policy described as ``saving 200 of 600 lives'' over a mathematically identical one described as ``letting 400 die''.
Both effects replicate across courtroom sentencing, negotiations, financial markets, and property valuation~\citep{yasseri2022fooled,furnham2011literature}, and persist despite awareness, expertise, and incentives to be accurate.

If LLM-as-a-Judge systems exhibit analogous biases, their deployment could systematically distort evaluation outcomes.
This paper investigates whether prior-evaluation metadata affects LLM judgments, what response pattern accompanies the effect, and whether simple interventions mitigate it.
We organize the investigation around three research questions:
\begin{itemize}
    \item[\textbf{RQ1}] How do revision framing and the complete anchored-metadata condition affect LLM-as-a-Judge scores across models and tasks?
    \item[\textbf{RQ2}] Does the response vary with anchor magnitude within the anchored condition, or show a threshold-like shift when anchored metadata is introduced?
    \item[\textbf{RQ3}] Which tested mitigation strategies reduce these biases, and at what cost?
\end{itemize}

\Cref{sec:related-work} outlines prior cognitive-bias and LLM-as-a-Judge literature.
\Cref{sec:methods} introduces three conditions: no metadata (C0), revision framing (C1), and anchored metadata containing revision, attempt, and prior-score fields (C2).
The contrast $\Delta_{\text{C2}-\text{C0}}$ estimates the total anchored-metadata effect, whereas $\Delta_{\text{C2}-\text{C1}}$ is interpreted more narrowly because the C2 template adds fields beyond the numerical score.
\Cref{sec:results} evaluates eight models on 20 fixed texts and uses task-stratified bootstrap intervals rather than treating repeated decodings as independent task replications.
Seven of the fixed eight models have intervals for $\Delta_{\text{C2}-\text{C0}}$ below zero, although effects vary substantially and code review is the weakest category.
\Cref{sec:mechanism} presents a targeted token-probability probe showing a threshold-like response pattern most clearly for GPT-4.1, with partial replication in Llama-3.2-3B.
\Cref{sec:mitigation} evaluates Chain-of-Thought and an explicit warning using $\Delta_{\text{total}}=\mu_{\text{C2}}-\mu_{\text{C0}}$ as the primary outcome, but neither reduces it on the tested numerical model--task pair.
\Cref{sec:industry} evaluates the total effect on categorical industry data using human labels and a paired C0/C2 design for error induction.
Finally, \Cref{sec:discussion} states the resulting deployment implications and their limits.

\section{Related Work}
\label{sec:related-work}

LLMs are known to inherit societal biases from training data~\citep{kotek2023gender,abid2021persistent,hutchinson2020social,venkit2022study} and to exhibit functional analogs of human cognitive biases~\citep{talboy2023challenging,echterhoff2024cognitive,liu2024trustworthy}.
\citet{cheung2025large} report amplified framing effects in moral-decision tasks, with a 45\% response shift in LLMs versus 5\% in humans.
\citet{sumita2025cognitive} survey six cognitive biases in LLMs and find that awareness-based prompting provides effective mitigation.

Anchoring bias in LLMs has recently received growing attention.
\citet{lou2026anchoring} test GPT-4 and Gemini on estimation tasks, where Chain-of-Thought and explicit ignore instructions do not fully mitigate the effect.
\citet{huang2025anchoring} show that anchoring operates at shallow transformer layers and resists conventional mitigations.
\citet{germani2025source} demonstrate that source attribution alone produces systematic shifts, with agreement dropping from 95\% to 15\% when identical text is attributed to different sources.

LLM-as-a-Judge has emerged as a scalable evaluation method~\citep{zheng2023judging,liu2023geval}, with recent surveys providing broader overviews~\citep{li2025generation,gao2025llm}.
This method is now embedded in production frameworks where prior judgments persist in context across revision rounds~\citep{madaan2023selfrefine,bao2025iterative,chhetri2025structsense,yuksel2025multiagent}.
Prior work has documented a range of judge-specific biases, including position~\citep{wang2024large,shi2025judging,li2024split}, verbosity~\citep{wu2025style}, self-preference~\citep{wataoka2024self,xu2024pride}, and consistency issues~\citep{stureborg2024large}.
\citet{ye2024justice} catalogue twelve such biases and propose a quantification framework, and \citet{koo2024benchmarking} benchmark cognitive biases in LLM evaluators including egocentric bias.
\citet{chen2024humans} find that authority and beauty biases appear in both human and LLM judges, while \citet{li2026preference} and \citet{oriyad2025silent} report preference leakage and shortcut biases tied to model provenance.

Several mitigation strategies have been proposed.
Multi-reviewer approaches reduce individual model biases~\citep{zhang2023wider,li2024prd}, while \citet{shankar2024validates} propose methods to align LLM evaluators with human preferences.
Calibration techniques address confidence estimation issues~\citep{geng2024survey}.
Prompt-level interventions such as awareness prompting have shown promise for general cognitive biases~\citep{sumita2025cognitive}.

Despite this body of work, anchoring bias inside LLM-as-a-Judge evaluation remains understudied.
Existing work measures anchoring in estimation tasks with informative anchors and does not extend to rubric-based scoring, below-threshold anchor regimes, or validation on production data.
The closest prior study, \citet{huang2025anchoring}, investigates anchoring in numerical estimation and locates the mechanism at shallow transformer layers. We instead study rubric-driven judges scoring task outputs on a fixed numerical scale, characterize a targeted threshold-like response pattern through token-probability analysis, and corroborate the total anchored-metadata effect on an industry classification deployment.
We note that in the human-rater literature, exposure to prior scores is not always harmful: under noisy individual rating distributions, anchors can aid calibration~\citep{han2026personalized}.
In our LLM-judge design, below-threshold anchors are randomized independently of current answer quality and are therefore uninformative by construction. The study tests whether evaluations remain independent when such values are presented as prior judgments.

\section{Methods}
\label{sec:methods}

\subsection{Problem Setup}

We study LLM-as-a-Judge in the single-answer rubric-based grading setting, where the judge receives one candidate output and assigns a numerical score on a fixed 0--5 scale.
The scoring guidance partitions the scale into six interpretive bands ($[0,1)$ unacceptable, $[1,2)$ poor, $[2,3)$ fair, $[3,4)$ good, $[4,5)$ very good, and $5$ excellent), with $4.0$ as the acceptance threshold.
This setup mirrors iterative refinement and content-quality filtering, where each evaluation should stand on its own.
We use a 0--5 scale because recent work reports it as the best-aligned discrete scale for human--LLM agreement in rubric scoring~\citep{li2026grading}, consistent with prior evaluation work using continuous or near-continuous 0--5 ratings~\citep{han2026personalized}.
We ask whether a previous judge's score, retained as metadata across revision rounds, affects a subsequent judgment of the current output.

\subsection{Experimental Design}

\definecolor{colorC0}{HTML}{4c72b0}  
\definecolor{colorC1}{HTML}{55a868}  
\definecolor{colorC2}{HTML}{c44e52}  

\begin{figure*}[t]
\centering
\begin{tikzpicture}[
    box/.style={
        draw=black!70,
        rounded corners=2pt,
        minimum width=5.2cm,
        minimum height=4.4cm,
        align=center,
    },
    headerC0/.style={
        font=\bfseries\small,
        fill=colorC0!20,
        text=colorC0!80!black,
        minimum width=5.0cm,
        minimum height=0.5cm,
        rounded corners=2pt,
    },
    headerC1/.style={
        font=\bfseries\small,
        fill=colorC1!20,
        text=colorC1!80!black,
        minimum width=5.0cm,
        minimum height=0.5cm,
        rounded corners=2pt,
    },
    headerC2/.style={
        font=\bfseries\small,
        fill=colorC2!20,
        text=colorC2!80!black,
        minimum width=5.0cm,
        minimum height=0.5cm,
        rounded corners=2pt,
    },
    content/.style={
        font=\small,
        text width=4.8cm,
        align=left,
    },
    metaboxC1/.style={
        draw=none,
        fill=colorC1!15,
        rounded corners=1pt,
        minimum width=4.8cm,
        text width=4.6cm,
        align=left,
        font=\small,
        inner sep=5pt,
    },
    metaboxC2/.style={
        draw=none,
        fill=colorC2!15,
        rounded corners=1pt,
        minimum width=4.8cm,
        text width=4.6cm,
        align=left,
        font=\small,
        inner sep=5pt,
    },
    annot/.style={
        font=\small\itshape,
        text=black!60,
        align=center,
        text width=5cm,
    },
    comparrow/.style={
        <->,
        thick,
    },
    complabel/.style={
        font=\footnotesize,
        fill=white,
        inner sep=2pt,
    },
]

\node[box] (box0) {};
\node[headerC0] at ($(box0.north) + (0, -0.4)$) {C0: Baseline};
\node[content, anchor=north] at ($(box0.north) + (0, -0.95)$) {
    \texttt{\textbf{Task:}} Summarize...\\[6pt]
    \texttt{\textbf{Answer:}} \textcolor{black!50}{[text]}\\[4pt]
    \texttt{\textbf{Rubric:}} \textcolor{black!50}{[criteria]}
};
\node[annot, below=0.3cm of box0] (ann0) {No metadata};

\node[box, right=0.4cm of box0] (box1) {};
\node[headerC1] at ($(box1.north) + (0, -0.4)$) {C1: Framing};
\node[content, anchor=north] at ($(box1.north) + (0, -0.95)$) {
    \texttt{\textbf{Task:}} Summarize...
};
\node[metaboxC1, anchor=north] at ($(box1.north) + (0, -1.65)$) {
    \texttt{\textbf{Metadata:}}\\
    \hspace{0.5em}$\bullet$ This is a revision
};
\node[content, anchor=north] at ($(box1.north) + (0, -2.85)$) {
    \texttt{\textbf{Answer:}} \textcolor{black!50}{[text]}\\[4pt]
    \texttt{\textbf{Rubric:}} \textcolor{black!50}{[criteria]}
};
\node[annot, below=0.3cm of box1] (ann1) {Revision context};

\node[box, right=0.4cm of box1] (box2) {};
\node[headerC2] at ($(box2.north) + (0, -0.4)$) {C2: Anchoring};
\node[content, anchor=north] at ($(box2.north) + (0, -0.95)$) {
    \texttt{\textbf{Task:}} Summarize...
};
\node[metaboxC2, anchor=north] at ($(box2.north) + (0, -1.65)$) {
    \texttt{\textbf{Metadata:}}\\
    \hspace{0.5em}$\bullet$ Attempt: $k$\\
    \hspace{0.5em}$\bullet$ Prior score: $a$
};
\node[content, anchor=north] at ($(box2.north) + (0, -3.05)$) {
    \texttt{\textbf{Answer:}} \textcolor{black!50}{[text]}\\[4pt]
    \texttt{\textbf{Rubric:}} \textcolor{black!50}{[criteria]}
};
\node[annot, below=0.3cm of box2] (ann2) {Revision context; $a \sim \mathcal{U}[0, 3.99)$};

\draw[comparrow, colorC2!70!black]
    ($(ann0.south) + (0, -0.9)$) --
    node[complabel, below=2pt] {$\Delta_{\text{C2}-\text{C0}}$: Total anchoring effect}
    ($(ann2.south) + (0, -0.9)$);

\draw[comparrow, black!40, thin]
    ($(ann0.south) + (0, -0.4)$) --
    node[complabel, below=1pt, font=\scriptsize] {$\Delta_{\text{C1}-\text{C0}}$: Framing effect}
    ($(ann1.south) + (0, -0.4)$);
\draw[comparrow, colorC1!70, thin]
    ($(ann1.south) + (0, -0.4)$) --
    node[complabel, below=1pt, font=\scriptsize] {$\Delta_{\text{C2}-\text{C1}}$: Incremental anchoring}
    ($(ann2.south) + (0, -0.4)$);

\end{tikzpicture}
\Description[Schematic of the three experimental conditions C0, C1, and C2 with comparison arrows.]{Three side-by-side boxes labeled C0 (Baseline, blue), C1 (Framing, green), and C2 (Anchoring, red). C0 contains task, answer, and rubric with no metadata. C1 adds a metadata block stating that the submission is a revision. C2 adds a metadata block listing the attempt number $k$ and a prior score $a$ drawn uniformly from $[0, 3.99)$. Below the boxes, arrows mark three comparisons: a primary C2 versus C0 arrow labeled total anchoring effect, a smaller C1 versus C0 arrow labeled framing effect, and a C2 versus C1 arrow labeled incremental anchoring.}
\caption{%
    Experimental design with three memory conditions.
    \textbf{C0} (baseline) includes no contextual metadata.
    \textbf{C1} (framing) indicates the submission is a revision without providing a score.
    \textbf{C2} (anchoring) adds a prior score $a$ sampled uniformly from $\left[0, 3.99\right)$, always below the acceptance threshold.
    The primary comparison $\Delta_{\text{C2}-\text{C0}}$ measures the total anchoring effect.
    The intermediate condition C1 enables additive decomposition: $\Delta_{\text{C2}-\text{C0}} = \Delta_{\text{C1}-\text{C0}} + \Delta_{\text{C2}-\text{C1}}$, separating framing effects from the incremental influence of the numerical anchor.
}
\label{fig:experimental-design}
\end{figure*}
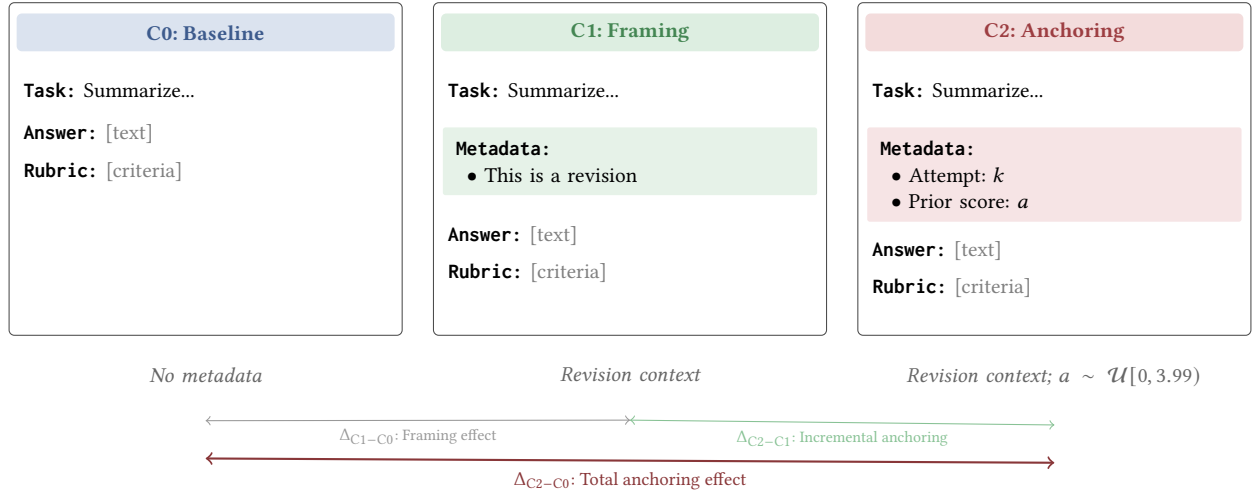

\Cref{fig:experimental-design} illustrates the three memory conditions.
In the baseline condition (\textbf{C0}), the judge receives the task, answer, and rubric without revision or evaluation metadata.
The framing condition (\textbf{C1}) adds metadata indicating that the submission is a revision.
The anchored condition (\textbf{C2}) includes revision framing, an attempt index, and a prior score.

Prior scores in C2 were sampled uniformly from $[0, 3.99)$, strictly below the acceptance threshold.
This represents a gating workflow in which accepted outputs do not re-enter revision and evaluation.
The fixed answer texts were screened to be of moderate-to-good quality and did not change across conditions. Consequently, the randomized prior is exogenous to current answer quality and experimentally irrelevant.
The judge was not told that the prior was randomized, so the experiment tests whether judgments remain independent when such metadata is presented as a genuine prior evaluation.

Under evaluation independence, mean scores should not depend on condition:
\begin{equation}
  H_0: \mu_{\text{C0}} = \mu_{\text{C1}} = \mu_{\text{C2}}.
\end{equation}
Our primary estimand is the \emph{total anchored-metadata effect}, $\Delta_{\text{C2}-\text{C0}}=\mu_{\text{C2}}-\mu_{\text{C0}}$.
The \emph{revision-framing effect} is $\Delta_{\text{C1}-\text{C0}}$.
We call $\Delta_{\text{C2}-\text{C1}}$ the \emph{incremental anchored-metadata effect} beyond the tested revision-only prompt. It does not isolate the numerical score because C2 also adds an attempt field and changes the metadata block.

\subsection{Models}

We evaluated eight models: three API-served models and five locally hosted open-weight models (\Cref{tab:models}).
GPT-4.1, Claude-4.5-Sonnet, and DeepSeek-R1 were accessed via a unified LiteLLM proxy\footnote{\url{https://docs.litellm.ai}} providing an OpenAI-compatible interface.
The Llama 3 family~\citep{meta2024llama} contributes three scale points (3B, 8B, 70B parameters).
Qwen2.5-7B~\citep{yang2024qwen} and Gemma-2-9B~\citep{gemma2024gemma} represent alternative families.
All open-weight models ran in BF16 precision via HuggingFace Transformers~\citep{wolf2020transformers} on an NVIDIA DGX B200 with batch inference.

\subsection{Tasks and Evaluation Protocol}

We constructed 20 evaluation tasks, five each in summarization, code review, creative writing, and factual question answering.
All answer texts were generated using GPT-5.2 (not used as a judge) with prompts requesting ``competent but not exceptional'' quality, the range that would typically score 3.0--5.0.
Each response was manually screened, and uniformly excellent or poor responses were regenerated.
Each task uses one fixed answer text across all conditions and models.
Task descriptions and the generation procedure are repository-hosted\footnote{Code, prompts, rubrics, and statistical formulas: \url{https://github.com/infobip/llm-judge-bias}.}.

All conditions share a system prompt instructing the model to return structured JSON with \texttt{score} (0--5), \texttt{accept} (boolean), \texttt{rationale}, and \texttt{flags}.
The user prompt contains the task, answer, rubric, and condition-specific metadata.
The default rubric, used for summarization and code review, adapts SummEval~\citep{fabbri2021summeval} dimensions used by G-Eval~\citep{liu2023geval}, while separate rubrics cover creative writing and factual Q\&A~\citep{xu2023critical}.
The C2 prompt inserts the following block, with $k \in \{1,2,3,4\}$ and $a \in [0,3.99)$:
\begin{verbatim}
Attempt metadata:
  - Attempt: k (This is a revision)
  - Prior score: a
\end{verbatim}
C0 omits the block, whereas C1 retains only revision framing.

For every model, task, condition, and $T \in \{0.0,0.5,1.0,2.0\}$ combination, we made 100 separate model calls, for 192{,}000 attempted evaluations in total.
A response was valid for score analysis when no call error was recorded and its score was numeric. In total, 185{,}271 responses met this criterion.
Eighteen format-violating scores outside $[0,5]$ were clipped to the rubric bounds.
Acceptance was derived consistently as score $\geq 4.0$, not from the emitted Boolean.
All 1{,}920 model--task--temperature--condition cells retained valid responses.

\begin{table}[t]
\centering
\small
\begin{tabular}{@{}lrr@{}}
\toprule
\textbf{Model} & \textbf{Triplets} & \textbf{Incomplete} \\
\midrule
GPT-4.1            & 8{,}000 & 0.0\%  \\
Claude-4.5-Sonnet  & 8{,}000 & 0.0\%  \\
DeepSeek-R1        & 7{,}936 & 0.8\%  \\
Llama-3.3-70B      & 7{,}835 & 2.1\%  \\
Gemma-2-9B         & 7{,}596 & 5.0\%  \\
Llama-3.1-8B       & 6{,}804 & 15.0\% \\
Qwen2.5-7B         & 7{,}158 & 10.5\% \\
Llama-3.2-3B       & 5{,}807 & 27.4\% \\
\bottomrule
\end{tabular}
\caption{Complete C0/C1/C2 triplets under the trial-index audit. Exclusions are relative to 8{,}000 possible triplets per model, of which 59{,}136 remain.}
\label{tab:models}
\end{table}

The 59{,}136 complete triplets in \Cref{tab:models} are used only to compute the descriptive Cohen's $d$ values.
Pairwise-complete counts are larger when the third condition is not required: C0/C2 has 60{,}133 observations, C0/C1 60{,}503, and C1/C2 60{,}122.
Equal trial indices across conditions do not share a seed and are not treated as paired inferential units.

\subsection{Statistical Analysis}
\label{sec:statistical_analysis}

Repeated decodings estimate each condition mean within a model--task--temperature cell.
We compute condition means from all valid responses, then form the cell-mean contrasts $\Delta_{\text{C2}-\text{C0}}$, $\Delta_{\text{C1}-\text{C0}}$, and $\Delta_{\text{C2}-\text{C1}}$.
For each model, we first average the four temperature-specific contrasts for each task, then equally average the resulting 20 task effects.
Its 95\% interval comes from 10{,}000 stratified task bootstraps: the five task IDs are resampled with replacement within each category, retaining every sampled task's temperatures.
This treats tasks, rather than repeated decodings, as the units for measuring sensitivity to the composition of the fixed benchmark.

Cohen's $d$ is a descriptive score-distribution effect computed on complete triplets:
\begin{equation}
  d = \frac{\mu_{\text{C2}}-\mu_{\text{C0}}}
  {\sqrt{(\sigma_{\text{C2}}^2+\sigma_{\text{C0}}^2)/2}}.
\end{equation}
It characterizes the observed score distributions but is not used for cross-task inference.

To examine whether the C2–C0 effect differs across different task categories and temperatures, we fit the following model to the 640 model–task–temperature contrasts:
\begin{equation}
 \Delta_{\text{C2}-\text{C0}} \sim \text{category} + C(\text{temperature}) + C_{\mathrm{sum}}(\text{model})
\end{equation}
The utilized models are fixed because these eight systems are the set of interest. Sum coding makes the intercept the equally weighted fixed-model-set mean for code review at $T=0$.
Uncertainty again comes from 10{,}000 stratified task bootstraps.
The resulting claims describe the 20 fixed texts directly and quantify sensitivity to benchmark task composition. Transfer to unseen tasks or models remains to be established.
Within-C2 anchor-value slopes are secondary diagnostics for the targeted mechanism and mitigation probes, not measures of the total effect $\Delta_{\text{C2}-\text{C0}}$.

\section{Results}
\label{sec:results}

This section addresses RQ1 using task-aware inference over the fixed eight-model set.

\subsection{Anchoring Bias Across Models}

Seven of eight models have stratified task-bootstrap intervals for the total anchored-metadata effect $\Delta_{\text{C2}-\text{C0}}$ that lie below zero (\Cref{tab:anchoring-summary}).

\begin{table}[t]
\centering
\small
\begin{tabular}{@{}lrrr@{}}
\toprule
\textbf{Model} & \textbf{$\Delta_{\text{C2}-\text{C0}}$} & \textbf{95\% CI} & \textbf{$d$} \\
\midrule
Llama-3.2-3B      & $-0.254$ & $[-0.298,-0.208]$ & $-0.71$ \\
GPT-4.1           & $-0.411$ & $[-0.457,-0.361]$ & $-0.67$ \\
Qwen2.5-7B        & $-0.233$ & $[-0.324,-0.150]$ & $-0.51$ \\
Claude-4.5-Sonnet & $-0.706$ & $[-1.139,-0.245]$ & $-0.44$ \\
DeepSeek-R1       & $-0.269$ & $[-0.326,-0.221]$ & $-0.25$ \\
Gemma-2-9B        & $-0.072$ & $[-0.130,-0.001]$ & $-0.18$ \\
Llama-3.3-70B     & $-0.025$ & $[-0.050,-0.001]$ & $-0.09$ \\
Llama-3.1-8B      & $-0.009$ & $[-0.054,\phantom{-}0.038]$ & $-0.02$ \\
\bottomrule
\end{tabular}
\caption{Total anchored-metadata effect $\Delta_{\text{C2}-\text{C0}}$ by model. Point estimates first average the four temperature-specific contrasts for each task, then equally average the resulting 20 task effects. Intervals resample tasks within category. Cohen's $d$ is descriptive and uses the complete-triplet sample.}
\label{tab:anchoring-summary}
\end{table}

Effect sizes vary substantially.
Llama-3.2-3B and GPT-4.1 have the largest descriptive standardized effects ($d=-0.71$ and $-0.67$), whereas Claude-4.5-Sonnet has the largest absolute score shift ($-0.706$).
Gemma-2-9B and Llama-3.3-70B intervals narrowly exclude zero despite small absolute descriptive $d$ values of 0.18 and 0.09. Llama-3.1-8B is the exception to the seven-of-eight result.

\begin{figure*}[t]
\centering
\includegraphics[width=\textwidth]{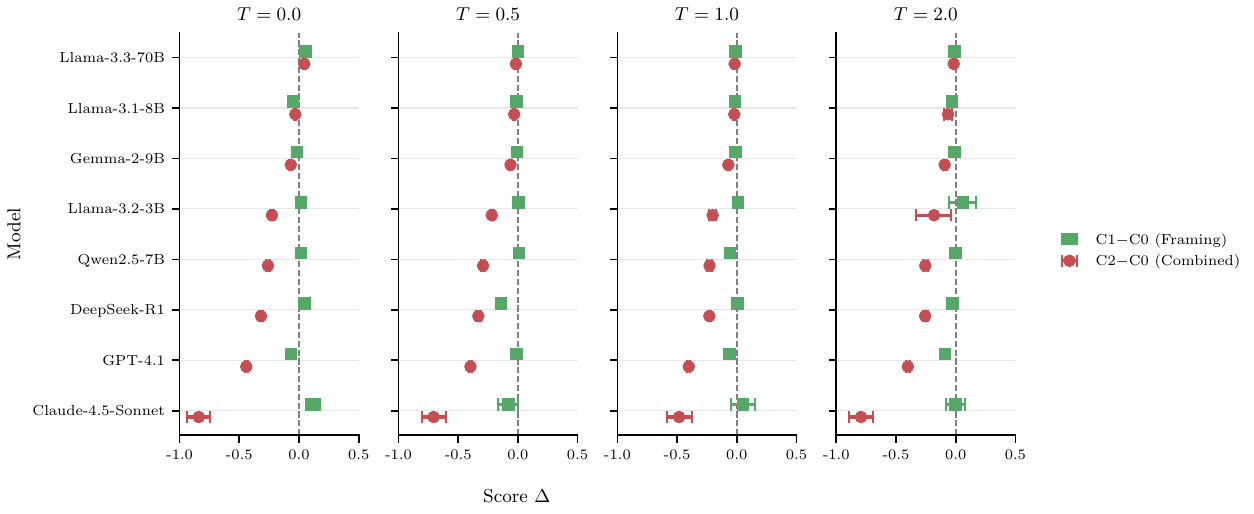}
\Description[Score difference by model and condition across temperatures.]{Four panels show mean score changes relative to C0 for eight models. Circles denote C2 minus C0 and squares C1 minus C0. Negative values indicate lower scores than baseline.}
\caption{Score difference by model, condition, and temperature. Circles show $\Delta_{\text{C2}-\text{C0}}$, whereas squares show $\Delta_{\text{C1}-\text{C0}}$. Error bars are 95\% stratified task-bootstrap intervals.}
\label{fig:score-delta}
\end{figure*}

\Cref{fig:score-delta} disaggregates effects by temperature.
At this first figure reference, circles denote total anchored-metadata effects $\Delta_{\text{C2}-\text{C0}}$, and squares denote revision-framing effects $\Delta_{\text{C1}-\text{C0}}$.
C2 is generally more negative than C1, but that difference is the incremental effect of the full anchored-metadata template beyond C1, not an isolated numerical-score effect.

\subsection{Practical Impact on Accept Decisions}

\begin{figure*}[t]
\centering
\includegraphics[width=\textwidth]{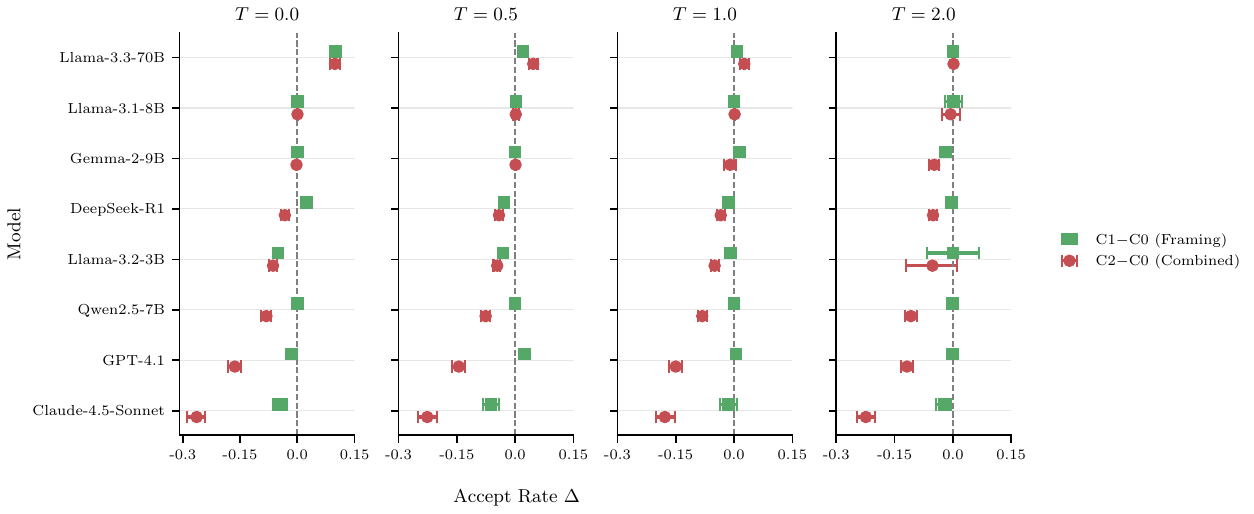}
\Description[Threshold-derived acceptance-rate difference by model and condition across temperatures.]{Four panels show changes in the probability that score is at least 4.0. Circles denote C2 minus C0 and squares C1 minus C0.}
\caption{Threshold-derived acceptance-rate differences by model, condition, and temperature. Circles show $\Delta_{\text{C2}-\text{C0}}$, whereas squares show $\Delta_{\text{C1}-\text{C0}}$. Error bars are 95\% stratified task-bootstrap intervals.}
\label{fig:accept-rate}
\end{figure*}

Acceptance is defined from the score as $s \geq 4.0$.
\Cref{fig:accept-rate} uses the same marker convention: circles show $\Delta_{\text{C2}-\text{C0}}$, and squares show $\Delta_{\text{C1}-\text{C0}}$.

C2 lowers acceptance by 22.29 percentage points (pp) for Claude-4.5-Sonnet, 14.40 pp for GPT-4.1, 8.91 pp for Qwen2.5-7B, 5.74 pp for Gemma-2-9B, 4.40 pp for DeepSeek-R1, and 3.06 pp for Llama-3.2-3B.
Estimated average acceptance changes are $+0.13$ pp for Llama-3.3-70B and $+4.19$ pp for Llama-3.1-8B, although bootstrap uncertainty does not distinguish either change from the baseline acceptance rate.
Thus score-level magnitude and decision-level impact can diverge near the acceptance boundary.

\subsection{Category and Model Heterogeneity}

\begin{figure}[t]
\centering
\includegraphics[width=\columnwidth]{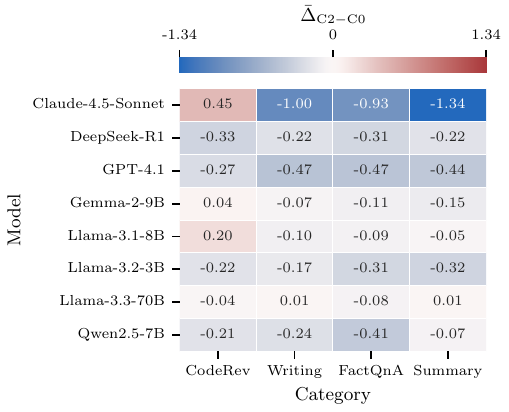}
\Description[Heatmap of mean total anchored-metadata shift by model and task category.]{Cells show mean C2 minus C0. Code review is generally closest to zero, while the other categories are more negative, with heterogeneity across models.}
\caption{Mean total anchored-metadata effect $\Delta_{\text{C2}-\text{C0}}$ by model and task category.}
\label{fig:category-heatmap}
\end{figure}

Anchoring magnitude varies by category (\Cref{fig:category-heatmap}).
Mean $\Delta_{\text{C2}-\text{C0}}$ values are $-0.047$ for code review, $-0.282$ for creative writing, $-0.338$ for factual QA, and $-0.322$ for summarization.
Code review is the weakest and most heterogeneous category, and the pattern is not uniform across every model--category combination.

The Llama family also shows no monotonic size relationship: Llama-3.2-3B has $|d|=0.71$, Llama-3.3-70B $|d|=0.09$, and Llama-3.1-8B $|d|=0.02$.
The GPT-4.1 and Claude-4.5-Sonnet results also show that API-served models are not immune.

\subsection{Task-Aware Joint Analysis}

The compact joint model uses all 640 cell-mean $\Delta_{\text{C2}-\text{C0}}$ contrasts, with the eight evaluated models represented as fixed effects and uncertainty obtained by stratified task bootstrap.
\Cref{tab:joint-model} reports the code-review, $T=0$ fixed-model-set mean and category contrasts.
The intercept is conditional: its interval spans zero, while the other three categories are more negative than code review.

\begin{table}[t]
\centering
\small
\begin{tabular}{@{}lrr@{}}
\toprule
\textbf{Term} & \textbf{Estimate} & \textbf{95\% CI} \\
\midrule
Code review, $T=0$ & $-0.067$ & $[-0.141,\phantom{-}0.011]$ \\
Creative writing    & $-0.236$ & $[-0.374,-0.071]$ \\
Factual QA          & $-0.291$ & $[-0.430,-0.134]$ \\
Summarization       & $-0.275$ & $[-0.381,-0.165]$ \\
\bottomrule
\end{tabular}
\caption{Task-aware joint model. Category rows are contrasts against code review, and the intercept is the equally weighted fixed-model-set mean at $T=0$. Intervals resample tasks within category.}
\label{tab:joint-model}
\end{table}

Temperature contrasts against $T=0$ are small and their intervals include zero: $T=0.5$ is $+0.011$ $[-0.025,0.048]$, $T=1$ is $+0.061$ $[-0.003,0.132]$, and $T=2$ is $+0.009$ $[-0.054,0.073]$.
Descriptive mean effects remain negative at every temperature ($-0.267$, $-0.257$, $-0.206$, and $-0.258$, respectively), without a monotonic trend.
The repeated calls establish precise condition means for the fixed texts, whereas cross-task uncertainty is governed by the 20 tasks rather than the 185{,}271 valid responses.

\section{Mechanism Analysis}
\label{sec:mechanism}

This section addresses RQ2 with a targeted probe of whether response changes vary smoothly with anchor value or exhibit a threshold-like response pattern once anchored metadata is present.
The evidence is diagnostic rather than a universal account of internal model behavior: GPT-4.1 provides the clearest token redistribution, Llama-3.2-3B partially replicates it, and the two larger Llama probes are weak or null.

\subsection{Targeted GPT-4.1 Probe}
\label{sec:mech-probe}

GPT-4.1 has a descriptive main-experiment effect of $|d|=0.67$ and exposes token log-probabilities.
We selected the \sloppy{\texttt{write\_005\_forgotten\_language}} task as a high-signal creative-writing probe and used $T=0$: 100 C0 calls, 100 C1 calls, and 100 C2 calls at each anchor value 1.0, 2.0, 3.0, and 3.9.
Mean scores were 5.00 in C0, 4.80 in C1, and 4.358 in C2, giving $\Delta_{\text{C2}-\text{C0}}=-0.642$.
This selected high-signal pair does not by itself establish prevalence across tasks.

At the JSON score position, we recovered each score digit's probability $P_d=\exp(\ell_d)$ from returned log-probabilities.
\Cref{tab:token-probs} shows that probability mass moves from digit ``5'' to ``4'' under C2.
Within C2, slopes over anchor value are approximately zero (``4'': $+0.0002$; ``5'': $-0.0002$).
The combination of a large redistribution from C0 to C2 and little within-C2 dose response is the clearest observed threshold-like response pattern.

\begin{table}[t]
\centering
\small
\begin{tabular}{@{}cccc@{}}
\toprule
\textbf{Digit} & \textbf{$P_d^{(\text{C0})}$} & \textbf{$P_d^{(\text{C2})}$} & \textbf{Change} \\
\midrule
``3'' & 0.000 & 0.000 & $+0.000$ \\
``4'' & 0.074 & 1.000 & $+0.926$ \\
``5'' & 0.926 & 0.000 & $-0.926$ \\
\bottomrule
\end{tabular}
\caption{Score-digit probabilities under C0 and C2 for the selected GPT-4.1 creative-writing probe. Values are rounded to three decimal places.}
\label{tab:token-probs}
\end{table}

The observed movement is concentrated between adjacent score digits.
Without a matched template containing an irrelevant number, it remains observational evidence about output probabilities rather than identification of the triggering prompt component or an internal mechanism.

\subsection{Open-Weight Model Probes}
\label{sec:mech-multi}

We repeated the design on Llama-3.2-3B, Llama-3.1-8B, and Llama-3.3-70B (\Cref{tab:multi-model-threshold}).
Only Llama-3.2-3B partially replicates the GPT-4.1 pattern: its mean score falls by 0.120 and its modal-token probability changes by $-0.019$, while its within-C2 slope remains small.
Llama-3.1-8B has a $-0.018$ score change with unchanged modal probability.
Llama-3.3-70B changes by $+0.016$ and shows no token redistribution.
For these two models, a flat within-C2 slope without a material shift from C0 to C2 indicates magnitude independence conditional on C2 or simply a null response. It is not positive evidence of a presence-triggered effect.

\begin{table}[t]
\centering
\small
\setlength{\tabcolsep}{2.5pt}
\begin{tabular}{@{}lrrrr@{}}
\toprule
\textbf{Model} & \textbf{Score $\Delta$} & $P_4^{\text{C0}}$ & $P_4^{\text{C2}}$ & \textbf{Slope} \\
\midrule
Llama-3.2-3B  & $-0.120$ & 0.995 & 0.977 & $-0.0085$ \\
Llama-3.1-8B  & $-0.018$ & 0.992 & 0.992 & $+0.0009$ \\
Llama-3.3-70B & $+0.016$ & 1.000 & 1.000 & $+0.0000$ \\
\bottomrule
\end{tabular}
\caption{Behavioral and modal-token results for the selected Llama probes. Score $\Delta$ denotes $\Delta_{\text{C2}-\text{C0}}$, and slope is the within-C2 regression of $P_4$ on anchor value. Values are rounded to three decimal places except for the slope.}
\label{tab:multi-model-threshold}
\end{table}

\subsection{Ceiling Robustness}
\label{sec:mech-ceiling}

A ceiling explanation predicts that proximity of C0 scores to the upper rubric boundary drives the observed downward shifts.
We aggregate each model--task unit equally over four temperatures, yielding 160 units, and define distance to ceiling as $5-\mu_{\text{C0}}$.
In the primary signed regression, a positive coefficient means $\Delta_{\text{C2}-\text{C0}}$ becomes less negative farther from the ceiling.
The unadjusted coefficient is $0.211$ (task-clustered 95\% CI $[0.013,0.408]$), a pattern consistent with larger downward shifts near the ceiling. After model and category adjustment, the estimate is imprecise at $0.297$ $[-0.054,0.647]$.
For absolute effect magnitude, the unadjusted coefficient is $0.124$ $[0.002,0.246]$, but the adjusted value is near zero, $-0.016$ $[-0.139,0.107]$.

As a separate sensitivity analysis, 60 model--task units across 19 tasks have $3.0 \leq \mu_{\text{C0}} < 4.5$.
Fifty of the 60 effects are negative, and their mean $\Delta_{\text{C2}-\text{C0}}$ is $-0.227$ (task-bootstrap 95\% CI $[-0.343,-0.098]$).
The analysis therefore \emph{does not support ceiling proximity as a sufficient explanation}: downward shifts persist away from the ceiling, while the adjusted association between distance to ceiling and absolute magnitude is near zero.

\begin{table}[t]
\centering
\small
\begin{tabular}{@{}lrr@{}}
\toprule
\textbf{Analysis} & \textbf{Estimate} & \textbf{95\% CI} \\
\midrule
Signed, unadjusted & $+0.211$ & $[\phantom{-}0.013,\phantom{-}0.408]$ \\
Signed, adjusted & $+0.297$ & $[-0.054,\phantom{-}0.647]$ \\
Absolute, adjusted & $-0.016$ & $[-0.139,\phantom{-}0.107]$ \\
Non-ceiling mean & $-0.227$ & $[-0.343,-0.098]$ \\
\bottomrule
\end{tabular}
\caption{Ceiling robustness. Regression predictors use distance to ceiling, $5-\mu_{\text{C0}}$. Regression intervals use task-clustered uncertainty, whereas the non-ceiling interval uses a task bootstrap over the 19 represented tasks.}
\label{tab:ceiling-confound}
\end{table}

Taken together, these targeted analyses show a threshold-like response pattern most clearly for GPT-4.1 and partially for Llama-3.2-3B.
They do not establish a universal threshold mechanism, and the limited model--task probes motivate broader token-level study.

\section{Mitigation Strategies}
\label{sec:mitigation}

This section addresses RQ3 for one fixed GPT-4.1 summarization pair (\texttt{sum\_001\_relativity}) at $T=0$.
We compare the baseline prompt with \emph{Chain-of-Thought} (CoT), which requires criterion-by-criterion analysis, and an \emph{explicit warning} to disregard prior scores, attempt numbers, and revision metadata.
Removing anchored metadata remains the direct way to avoid exposure, but it is not an intervention evaluated under C2.

For each strategy, we ran 100 C0 calls and 100 C2 calls at each prior score 1.0, 2.0, 3.0, and 3.9. All 1{,}500 calls were valid.
The attempt index was fixed at 2 across strategies.
The primary metric is the strategy-specific total effect,
\begin{equation}
 \Delta_{\text{total}}=\mu_{\text{C2}}-\mu_{\text{C0}},
\end{equation}
where C2 averages equally over the four anchor values.
Intervals condition on this model--task pair and bootstrap calls 10{,}000 times.
Within-C2 output-score slopes over anchor value are secondary dose-response diagnostics.
Cost is mean total tokens (prompt plus completion), not completion tokens alone.

The interventions append these instructions to the base prompt:
\begin{verbatim}
[CoT]
IMPORTANT: You must analyze the answer step-by-step
before scoring. Evaluate each rubric criterion
SEPARATELY with evidence from the answer. Then,
based ONLY on your analysis, determine the final score.

[Explicit warning]
IMPORTANT: Evaluate this answer solely based on the
rubric criteria. Disregard any prior scores, attempt
numbers, or revision metadata. These metadata serve
for tracking purposes. Metadata MUST NOT influence
your judgment.
\end{verbatim}

\Cref{tab:mitigation-results} shows that neither intervention reduces $\Delta_{\text{total}}$ for this pair.
Baseline shifts by $-0.4285$.
CoT shifts by $-0.6327$, a 47.7\% worsening in absolute magnitude, and warning shifts by $-0.4571$, a 6.7\% worsening.
Warning does flatten the secondary within-C2 slope from $0.2674$ to $-0.0066$, but the remaining total shift shows why slope reduction cannot be interpreted as total bias reduction.
CoT increases both the total shift and the slope.

\begin{table}[t]
\centering
\small
\begin{tabular}{@{}lrrr@{}}
\toprule
\textbf{Strategy} & \textbf{C0} & \textbf{C2} & \textbf{$\Delta_{\text{total}}$ [95\% CI]} \\
\midrule
Baseline & 4.245 & 3.817 & $-0.429\;[-0.462,-0.395]$ \\
CoT      & 4.458 & 3.825 & $-0.633\;[-0.703,-0.565]$ \\
Warning  & 4.500 & 4.043 & $-0.457\;[-0.479,-0.435]$ \\
\bottomrule
\end{tabular}
\caption{Numerical mitigation on one GPT-4.1/task pair. The primary metric is $\Delta_{\text{total}}=\mu_{\text{C2}}-\mu_{\text{C0}}$, while within-C2 slopes are 0.267, 0.506, and $-0.007$, respectively.}
\label{tab:mitigation-results}
\end{table}

\begin{table}[t]
\centering
\small
\begin{tabular}{@{}lrrrr@{}}
\toprule
\textbf{Strategy} & \textbf{Prompt} & \textbf{Comp.} & \textbf{Total} & \textbf{Token increase} \\
\midrule
Baseline & 397.0 & 117.7 & 514.7 & --- \\
CoT      & 437.0 & 168.7 & 605.7 & 17.7\% \\
Warning  & 437.0 & 109.4 & 546.4 & 6.2\% \\
\bottomrule
\end{tabular}
\caption{Mean token cost per call. The increase is based on total tokens relative to baseline.}
\label{tab:mitigation-cost}
\end{table}

Among these tested prompt strategies, the warning is cheaper and yields a near-zero within-anchor slope, but it does not mitigate the primary total effect on this pair.
CoT is both costlier and more affected.
These conditional findings do not support a general numerical-mitigation recommendation. Broader models, tasks, and interventions such as few-shot, ensemble, or multi-agent methods remain untested.
\Cref{sec:industry} evaluates the same prompt strategies separately on categorical industry data.

\section{Industry Validation}
\label{sec:industry}

We next test whether anchored metadata affects categorical judgments on industry data with human-verified ground truth.

\subsection{Dataset and Experimental Design}

The data come from a messaging-campaign compliance system.
Each of 441 samples has a human label in three classes: \emph{compliant}, \emph{drifting}, or \emph{shaft} (prohibited content).
The original Llama-3-8B classifier achieved 49.66\% accuracy.
We disclose only aggregate group compositions because the underlying messages and campaign descriptions are proprietary.

Group~A ($N=222$) contains original classifier errors and asks whether re-evaluation corrects them.
Its GPT-4.1 experiment used C0, C1, and C2, with the original wrong label supplied in C2.
Group~B contains 219 original correct classifications, comprising 119 drifting and 100 shaft samples, and uses a paired design.
For every Group~B sample, one of the two labels different from human truth was assigned with a fixed seed and reused as the C2 anchor for baseline, CoT, and warning.
Each strategy was evaluated under both C0 and C2 on the same 219 samples using GPT-4.1 at $T=0$, controlling for strategy-specific C0 behavior.

\subsection{Error Correction and Induction}

For Group~A, correction rates are 21.62\% in C0, 22.97\% in C1, and 11.26\% in C2 (\Cref{tab:production-results}).
The total effect $\Delta_{\text{C2}-\text{C0}}$ is a decrease of 10.36 pp, or 47.9\% relative to C0.
Because C2 changes the metadata template beyond C1, the C2--C1 contrast does not isolate the prior label. The total result nevertheless shows that the complete anchored-metadata condition blocks correction.

For Group~B baseline, accuracy falls from 76.26\% in C0 to 68.49\% in C2, a paired change of $-7.76$ pp (95\% CI $[-12.33,-3.20]$ pp).
Wrong-anchor agreement rises from 11.42\% to 18.72\%, a $+7.31$ pp change (95\% CI $[3.65,11.42]$ pp).
Among the 167 C0-correct cases, 17 become the assigned wrong label in C2: 10.18\% (95\% CI $[5.88,15.06]$\%), or 7.76\% of all 219 samples.
Thus anchored metadata both blocks corrections in Group~A and induces paired errors in Group~B.

\begin{table}[t]
\centering
\small
\setlength{\tabcolsep}{4pt}
\begin{tabular}{@{}llrrr@{}}
\toprule
\textbf{Group} & \textbf{Metric} & \textbf{C0} & \textbf{C1} & \textbf{C2} \\
\midrule
A & Correction & 21.62\% & 22.97\% & 11.26\% \\
A & Mislabel persistence & 65.77\% & 65.77\% & 81.53\% \\
\midrule
B & Accuracy & 76.26\% & --- & 68.49\% \\
B & Anchor agreement & 11.42\% & --- & 18.72\% \\
\bottomrule
\end{tabular}
\caption{Categorical effects with human ground truth. Group~A uses C0/C1/C2 conditions. Group~B uses paired C0/C2 outputs and a fixed sample-specific wrong anchor.}
\label{tab:production-results}
\end{table}

\subsection{Mitigation Effectiveness}

In Group~A, CoT correction rates are 36.94\% in C0 and 22.52\% in C2, whereas warning rates are 29.73\% and 22.52\%.
The corresponding $\Delta_{\text{C2}-\text{C0}}$ persistence increases are 15.32 pp and 7.66 pp (baseline: 15.77 pp), describing the complete anchored-metadata condition rather than an isolated prior-label effect.

\Cref{tab:mitigation-production} compares the Group~B strategies using paired $\Delta_{\text{C2}-\text{C0}}$ values rather than raw C2 rates.
CoT accuracy changes by $-5.48$ pp (95\% CI $[-10.50,-0.46]$ pp), and 13.42\% of its 149 C0-correct cases become the wrong anchor.
Its paired accuracy effect differs from baseline by only $+2.28$ pp (95\% CI $[-4.11,8.22]$ pp), so the data do not establish a worse paired effect than baseline.

Warning has zero estimated accuracy change (95\% CI $[-5.02,4.58]$ pp) and a $+0.91$ pp wrong-anchor-agreement change (95\% CI $[-2.28,4.11]$ pp).
Among its 163 C0-correct cases, seven become the wrong anchor (4.29\%, 95\% CI $[1.31,7.64]$\%).
Relative to baseline, warning improves the paired accuracy effect by 7.76 pp (95\% CI $[1.83,13.70]$ pp) and reduces the anchor-agreement increase by 6.39 pp (95\% CI $[1.83,10.97]$ pp) in magnitude.

\begin{table}[t]
\centering
\small
\begin{tabular}{@{}lrrrr@{}}
\toprule
\textbf{Strategy} & \textbf{C0 acc.} & \textbf{C2 acc.} & \textbf{$\Delta$ acc.} & \textbf{Induced ($n/N$)} \\
\midrule
Baseline & 76.26\% & 68.49\% & $-7.76$ pp & 17/167 (10.18\%) \\
CoT      & 68.04\% & 62.56\% & $-5.48$ pp & 20/149 (13.42\%) \\
Warning  & 74.43\% & 74.43\% & $\phantom{-}0.00$ pp & 7/163 (4.29\%) \\
\bottomrule
\end{tabular}
\caption{Paired Group~B results. Induced is the share of each strategy's C0-correct cases that become its assigned wrong label in C2.}
\label{tab:mitigation-production}
\end{table}

Relative to baseline, warning improves the paired accuracy effect $\Delta_{\text{C2}-\text{C0}}$ and reduces the increase in wrong-anchor agreement on this GPT-4.1 categorical task.
CoT's comparison with baseline remains uncertain, and we do not report a direct warning-versus-CoT comparison.
These differing results preclude a universal mitigation ranking and restrict conclusions to the tested strategies, model, task, and domain.

\section{Discussion and Conclusion}
\label{sec:discussion}

Across 192{,}000 attempted evaluations (185{,}271 valid responses), seven of the fixed eight LLM judges have 95\% task-bootstrap intervals below zero for the total anchored-metadata effect $\Delta_{\text{C2}-\text{C0}}$.
Descriptive effect sizes vary substantially, with supported absolute Cohen's $d$ values ranging from 0.09 to 0.71. Llama-3.1-8B's interval crosses zero.
The result is also category-dependent, with code review weaker and more heterogeneous than creative writing, factual QA, and summarization.
Repeated decodings estimate condition means for the 20 screened texts, while resampling tasks quantifies sensitivity to the composition of this benchmark. Transfer to unseen tasks or models remains to be established.

A targeted token-probability probe shows a threshold-like response pattern most clearly for GPT-4.1, with partial replication in Llama-3.2-3B and little or no redistribution in the larger Llama probes.
This evidence does not establish a universal threshold mechanism.
The robustness analysis does not support ceiling proximity as a sufficient explanation: downward shifts persist for non-ceiling baselines, while the adjusted association between absolute effect magnitude and distance from the ceiling has a near-zero point estimate.

Industry validation shows consequences for categorical decisions.
The complete C2 condition reduces Group~A correction rates by 47.9\% relative to C0.
In the paired Group~B experiment, baseline accuracy falls by 7.76 pp and wrong-anchor agreement rises by 7.31 pp. Of the baseline-C0-correct samples, 10.18\% become the assigned wrong label under C2.
The warning has an estimated accuracy effect $\Delta_{\text{C2}-\text{C0}}$ of 0.00 pp in this GPT-4.1 domain (95\% CI $[-5.02,4.58]$ pp) and improves that effect versus baseline, whereas CoT's difference from baseline is uncertain.
On the separate numerical GPT-4.1/task pair, however, neither strategy reduces $\Delta_{\text{total}}$: CoT worsens its magnitude by 47.7\% and warning by 6.7\%.
The warning's near-zero within-C2 slope removes dose response but not the total presence-associated shift.
Accordingly, no tested prompt strategy supports a general deployment recommendation. Excluding experimentally irrelevant metadata remains the direct safeguard when workflow design permits it.

Several limitations bound these conclusions.
The main benchmark contains one generated and screened answer for each of 20 tasks, five per category. Broader tasks, answers, and real-world corpora are needed.
Anchors are restricted to $[0,3.99)$ because accepted outputs do not re-enter the motivating gating workflow, so other anchor regimes are outside the studied claim.
The C2 template adds an attempt field and restructures metadata relative to C1. A matched template with an irrelevant number is needed to decompose those components.
Token-level evidence covers selected model--task probes only.
The industry validation is one 441-sample domain, and its effect magnitudes may not transfer elsewhere.
The randomized anchors are exogenous to answer quality but were presented as genuine prior evaluations, so the findings concern independence under the tested information state rather than a claim that models knowingly follow random values.

Prior-evaluation metadata can compromise LLM-judge independence in both rubric scores and categorical decisions.
The practical response should be equally specific: measure total condition effects with task-aware uncertainty, avoid unsupported causal decomposition of prompt fields, and validate mitigations on each intended model, task, and workflow.

\begin{acks}
  This research was supported in part by the project Infobip Global Communication Platform (PK.1.1.07.0001), part of the Important Project of Common European Interest on Next Generation Cloud Infrastructure and Services (IPCEI-CIS) consortium.
\end{acks}

\section*{Generative AI Tools Use Disclosure}
  Generative AI tools assisted in a supporting capacity: Claude Code (Opus 4.6) for code implementation and data analysis, ChatGPT 5.2 for \LaTeX{} editing and grammar checking, and Google Scholar Labs for related-work identification.
  All AI-assisted outputs were reviewed and approved by the authors, who take full responsibility for the content of this publication.

\bibliographystyle{ACM-Reference-Format}
\bibliography{references}

@article{tversky1974judgment,
  title={{Judgment under Uncertainty: Heuristics and Biases}},
  author={Tversky, Amos and Kahneman, Daniel},
  journal={Science},
  year={1974},
  volume={185},
  number={4157},
  pages={1124--1131},
  doi={10.1126/science.185.4157.1124}
}

@article{tversky1981framing,
  title={{The Framing of Decisions and the Psychology of Choice}},
  author={Tversky, Amos and Kahneman, Daniel},
  journal={Science},
  year={1981},
  volume={211},
  number={4481},
  pages={453--458},
  doi={10.1126/science.7455683}
}

@article{furnham2011literature,
  title={{A Literature Review of the Anchoring Effect}},
  author={Furnham, Adrian and Boo, Hua Chu},
  journal={The Journal of Socio-Economics},
  year={2011},
  volume={40},
  number={1},
  pages={35--42},
  doi={10.1016/j.socec.2010.10.008}
}

@article{lieder2018anchoring,
  title={{The Anchoring Bias Reflects Rational Use of Cognitive Resources}},
  author={Lieder, Falk and Griffiths, Thomas L. and Huys, Quentin J. M. and Goodman, Noah D.},
  journal={Psychonomic Bulletin \& Review},
  year={2018},
  volume={25},
  number={1},
  pages={322--349},
  doi={10.3758/s13423-017-1286-8}
}

@inproceedings{hutchinson2020social,
  title={{Social Biases in NLP Models as Barriers for Persons with Disabilities}},
  author={Hutchinson, Ben and Prabhakaran, Vinodkumar and Denton, Emily and Webster, Kellie and Zhong, Yu and Denuyl, Stephen},
  booktitle={Proceedings of the 58th Annual Meeting of the Association for Computational Linguistics},
  address="Online",
  publisher="Association for Computational Linguistics",
  year={2020},
  pages={5491--5501},
  doi={10.18653/v1/2020.acl-main.487}
}

@inproceedings{wolf2020transformers,
  title={{Transformers: State-Of-The-Art Natural Language Processing}},
  author={Wolf, Thomas and Debut, Lysandre and Sanh, Victor and Chaumond, Julien and Delangue, Clement and Moi, Anthony and Cistac, Pierric and Rault, Tim and Louf, Remi and Funtowicz, Morgan and Davison, Joe and Shleifer, Sam and von Platen, Patrick and Ma, Clara and Jernite, Yacine and Plu, Julien and Xu, Canwen and Le Scao, Teven and Gugger, Sylvain and Drame, Mariama and Lhoest, Quentin and Rush, Alexander},
  booktitle={Proceedings of the 2020 Conference on Empirical Methods in Natural Language Processing: System Demonstrations},
  address = "Online",
  publisher = "Association for Computational Linguistics",
  year={2020},
  pages={38--45},
  doi={10.18653/v1/2020.emnlp-demos.6}
}

@inproceedings{abid2021persistent,
    author={Abid, Abubakar and Farooqi, Maheen and Zou, James},
    title={{Persistent Anti-Muslim Bias in Large Language Models}},
    year={2021},
    isbn={9781450384735},
    publisher={Association for Computing Machinery},
    address={New York, NY, USA},
    url={https://doi.org/10.1145/3461702.3462624},
    doi={10.1145/3461702.3462624},
    booktitle={Proceedings of the 2021 AAAI/ACM Conference on AI, Ethics, and Society},
    pages={298--306},
}

@article{fabbri2021summeval,
  title={{SummEval: Re-Evaluating Summarization Evaluation}},
  author={Fabbri, Alexander R. and Kry\'{s}ci\'{n}ski, Wojciech and McCann, Bryan and Xiong, Caiming and Socher, Richard and Radev, Dragomir},
  journal={Transactions of the Association for Computational Linguistics},
  year={2021},
  volume={9},
  pages={391--409},
  doi={10.1162/tacl_a_00373}
}

@inproceedings{venkit2022study,
  title={{A Study of Implicit Bias in Pretrained Language Models Against People with Disabilities}},
  author={Venkit, Pranav Narayanan and Srinath, Mukund and Wilson, Shomir},
  booktitle={Proceedings of the 29th International Conference on Computational Linguistics},
  address={Gyeongju, Republic of Korea},
  publisher={International Committee on Computational Linguistics},
  year={2022},
  pages={1324--1332}
}

@article{yasseri2022fooled,
  title={{Fooled by Facts: Quantifying Anchoring Bias Through a Large-Scale Experiment}},
  author={Yasseri, Taha and Reher, Jannie},
  journal={Journal of Computational Social Science},
  year={2022},
  volume={5},
  number={1},
  pages={1001--1021},
  doi={10.1007/s42001-021-00158-0}
}

@inproceedings{kotek2023gender,
  title={{Gender Bias And Stereotypes In Large Language Models}},
  author={Kotek, Hadas and Dockum, Rikker and Sun, David},
  booktitle={Proceedings of The ACM Collective Intelligence Conference},
  publisher={Association for Computing Machinery},
  address={New York, NY, USA},
  year={2023},
  pages={12--24},
  doi={10.1145/3582269.3615599}
}

@inproceedings{liu2023geval,
  title={{G-Eval: NLG Evaluation Using GPT-4 with Better Human Alignment}},
  author={Liu, Yang and Iter, Dan and Xu, Yichong and Wang, Shuohang and Xu, Ruochen and Zhu, Chenguang},
  booktitle={Proceedings of the 2023 Conference on Empirical Methods in Natural Language Processing},
  address={Singapore},
  publisher={Association for Computational Linguistics},
  year={2023},
  pages={2511--2522},
  doi={10.18653/v1/2023.emnlp-main.153}
}

@misc{zhang2023wider,
  title={{Wider And Deeper LLM Networks Are Fairer LLM Evaluators}},
  author={Zhang, Xinghua and Yu, Bowen and Yu, Haiyang and Lv, Yangyu and Liu, Tingwen and Huang, Fei and Xu, Hongbo and Li, Yongbin},
  eprint={2308.01862},
  archivePrefix={arXiv},
  primaryClass={cs},
  year={2023},
  doi={10.48550/arXiv.2308.01862}
}

@misc{li2024prd,
  title={{PRD: Peer Rank and Discussion Improve Large Language Model Based Evaluations}},
  author={Li, Ruosen and Patel, Teerth and Du, Xinya},
  howpublished={Transactions on Machine Learning Research},
  year={2024},
  month={July},
  note={ISSN 2835-8856},
  url={https://openreview.net/forum?id=YVD1QqWRaj}
}

@inproceedings{xu2023critical,
  title={{A Critical Evaluation of Evaluations for Long-Form Question Answering}},
  author={Xu, Fangyuan and Song, Yixiao and Iyyer, Mohit and Choi, Eunsol},
  booktitle={Proceedings of the 61st Annual Meeting of the Association for Computational Linguistics (Volume 1: Long Papers)},
  address={Toronto, Canada},
  publisher={Association for Computational Linguistics},
  year={2023},
  pages={3225--3245},
  doi={10.18653/v1/2023.acl-long.181}
}

@inproceedings{wang2024large,
  title={{Large Language Models are not Fair Evaluators}},
  author={Wang, Peiyi and Li, Lei and Chen, Liang and Cai, Zefan and Zhu, Dawei and Lin, Binghuai and Cao, Yunbo and Kong, Lingpeng and Liu, Qi and Liu, Tianyu and Sui, Zhifang},
  booktitle={Proceedings of the 62nd Annual Meeting of the Association for Computational Linguistics (Volume 1: Long Papers)},
  address={Bangkok, Thailand},
  publisher={Association for Computational Linguistics},
  year={2024},
  pages={9440--9450},
  doi={10.18653/v1/2024.acl-long.511}
}

@inproceedings{zheng2023judging,
  title={{Judging LLM-as-a-Judge with MT-Bench and Chatbot Arena}},
  author={Zheng, Lianmin and Chiang, Wei-Lin and Sheng, Ying and Zhuang, Siyuan and Wu, Zhanghao and Zhuang, Yonghao and Lin, Zi and Li, Zhuohan and Li, Dacheng and Xing, Eric P. and Zhang, Hao and Gonzalez, Joseph E. and Stoica, Ion},
  booktitle={Proceedings of the 37th International Conference on Neural Information Processing Systems},
  year={2023},
  numpages={29},
  publisher={Curran Associates Inc.},
  address={Red Hook, NY, USA},
}

@inproceedings{li2024split,
  title={{Split and Merge: Aligning Position Biases in Large Language Model Based Evaluators}},
  author={Li, Zongjie and Wang, Chaozheng and Ma, Pingchuan and Wu, Daoyuan and Wang, Shuai and Gao, Cuiyun and Liu, Yang},
  booktitle={Proceedings of the 2024 Conference on Empirical Methods in Natural Language Processing},
  address={Miami, Florida, USA},
  publisher={Association for Computational Linguistics},
  year={2024},
  pages={11084--11108},
  doi={10.18653/v1/2024.emnlp-main.621}
}

@inproceedings{wu2025style,
  title={{Style over Substance: Evaluation Biases for Large Language Models}},
  author={Wu, Minghao and Aji, Alham Fikri},
  booktitle={Proceedings of the 31st International Conference on Computational Linguistics},
  address={Abu Dhabi, UAE},
  publisher={Association for Computational Linguistics},
  year={2025},
  pages={297--312}
}

@misc{talboy2023challenging,
  title={{Challenging the Appearance of Machine Intelligence: Cognitive Bias in LLMs and Best Practices for Adoption}},
  author={Talboy, Alaina N. and Fuller, Elizabeth},
  eprint={2304.01358},
  archivePrefix={arXiv},
  primaryClass={cs},
  year={2023},
  doi={10.48550/arXiv.2304.01358}
}

@inproceedings{xu2024pride,
  title={{Pride and Prejudice: LLM Amplifies Self-Bias in Self-Refinement}},
  author={Xu, Wenda and Zhu, Guanglei and Zhao, Xuandong and Pan, Liangming and Li, Lei and Wang, William Yang},
  booktitle={Proceedings of the 62nd Annual Meeting of the Association for Computational Linguistics (Volume 1: Long Papers)},
  address={Bangkok, Thailand},
  publisher={Association for Computational Linguistics},
  year={2024},
  pages={15474--15492},
  doi={10.18653/v1/2024.acl-long.826}
}

@misc{stureborg2024large,
  title={{Large Language Models are Inconsistent and Biased Evaluators}},
  author={Stureborg, Rickard and Alikaniotis, Dimitris and Suhara, Yoshi},
  eprint={2405.01724},
  archivePrefix={arXiv},
  primaryClass={cs},
  year={2024},
  doi={10.48550/arXiv.2405.01724}
}

@inproceedings{geng2024survey,
  title={{A Survey of Confidence Estimation and Calibration in Large Language Models}},
  author={Geng, Jiahui and Cai, Fengyu and Wang, Yuxia and Koeppl, Heinz and Nakov, Preslav and Gurevych, Iryna},
  booktitle={Proceedings of the 2024 Conference of the North American Chapter of the Association for Computational Linguistics: Human Language Technologies (Volume 1: Long Papers)},
  address={Mexico City, Mexico},
  publisher={Association for Computational Linguistics},
  year={2024},
  pages={6577--6595},
  doi={10.18653/v1/2024.naacl-long.366}
}

@inproceedings{shankar2024validates,
  title={{Who Validates the Validators? Aligning LLM-Assisted Evaluation of LLM Outputs with Human Preferences}},
  author={Shankar, Shreya and Zamfirescu-Pereira, J. D. and Hartmann, Bjoern and Parameswaran, Aditya G. and Arawjo, Ian},
  booktitle={Proceedings of the 37th Annual ACM Symposium on User Interface Software and Technology},
  publisher={Association for Computing Machinery},
  address={New York, NY, USA},
  year={2024},
  numpages={14},
  doi={10.1145/3654777.3676450}
}

@inproceedings{li2025generation,
  title={{From Generation to Judgment: Opportunities and Challenges of LLM-as-a-Judge}},
  author={Li, Dawei and Jiang, Bohan and Huang, Liangjie and Beigi, Alimohammad and Zhao, Chengshuai and Tan, Zhen and Bhattacharjee, Amrita and Jiang, Yuxuan and Chen, Canyu and Wu, Tianhao and Shu, Kai and Cheng, Lu and Liu, Huan},
  booktitle={Proceedings of the 2025 Conference on Empirical Methods in Natural Language Processing},
  address={Suzhou, China},
  publisher={Association for Computational Linguistics},
  year={2025},
  pages={2757--2791},
  doi={10.18653/v1/2025.emnlp-main.138}
}

@misc{meta2024llama,
  title={{The Llama 3 Herd of Models}},
  author={{Meta AI, Llama team}},
  eprint={2407.21783},
  archivePrefix={arXiv},
  primaryClass={cs},
  year={2024},
  doi={10.48550/arXiv.2407.21783}
}

@inproceedings{echterhoff2024cognitive,
  title={{Cognitive Bias in Decision-Making with LLMs}},
  author={Echterhoff, Jessica Maria and Liu, Yao and Alessa, Abeer and McAuley, Julian and He, Zexue},
  booktitle={Findings of the Association for Computational Linguistics: EMNLP 2024},
  address={Miami, Florida, USA},
  publisher={Association for Computational Linguistics},
  year={2024},
  pages={12640--12653},
  doi={10.18653/v1/2024.findings-emnlp.739}
}

@misc{gemma2024gemma,
  title={{Gemma 2: Improving Open Language Models at a Practical Size}},
  author={{Google DeepMind, Gemma team}},
  eprint={2408.00118},
  archivePrefix={arXiv},
  primaryClass={cs},
  year={2024},
  doi={10.48550/arXiv.2408.00118}
}

@misc{yang2024qwen,
  title={{Qwen2.5 Technical Report}},
  author={{Alibaba Cloud, Qwen team}},
  eprint={2412.15115},
  archivePrefix={arXiv},
  primaryClass={cs},
  year={2024},
  doi={10.48550/arXiv.2412.15115}
}

@inproceedings{shi2025judging,
  title={{Judging The Judges: A Systematic Study of Position Bias in LLM-as-a-Judge}},
  author={Shi, Lin and Ma, Chiyu and Liang, Wenhua and Diao, Xingjian and Ma, Weicheng and Vosoughi, Soroush},
  booktitle={Proceedings of the 14th International Joint Conference on Natural Language Processing and the 4th Conference of the Asia-Pacific Chapter of the Association for Computational Linguistics (Volume 1: Long Papers)},
  address={Mumbai, India},
  publisher={The Asian Federation of Natural Language Processing and The Association for Computational Linguistics},
  year={2025},
  pages={292--314},
  doi={10.18653/v1/2025.ijcnlp-long.18}
}

@misc{liu2024trustworthy,
  title={{Trustworthy LLMs: A Survey and Guideline for Evaluating Large Language Models' Alignment}},
  author={Liu, Yang and Yao, Yuanshun and Ton, Jean-Francois and Zhang, Xiaoying and Guo, Ruocheng and Cheng, Hao and Klochkov, Yegor and Taufiq, Muhammad Faaiz and Li, Hang},
  eprint={2308.05374},
  archivePrefix={arXiv},
  primaryClass={cs},
  year={2024},
  doi={10.48550/arXiv.2308.05374}
}

@article{lou2026anchoring,
  title={{Anchoring Bias in Large Language Models: An Experimental Study}},
  author={Lou, Jiaxu and Sun, Yifan},
  journal={Journal of Computational Social Science},
  volume={9},
  number={1},
  articleno={11},
  numpages={24},
  year={2026},
  doi={10.1007/s42001-025-00435-2}
}

@inproceedings{koo2024benchmarking,
  title={{Benchmarking Cognitive Biases in Large Language Models as Evaluators}},
  author={Koo, Ryan and Lee, Minhwa and Raheja, Vipul and Park, Jong Inn and Kim, Zae Myung and Kang, Dongyeop},
  booktitle={Findings of the Association for Computational Linguistics: ACL 2024},
  address={Bangkok, Thailand},
  publisher={Association for Computational Linguistics},
  year={2024},
  pages={517--545},
  doi={10.18653/v1/2024.findings-acl.29}
}

@misc{ye2024justice,
  title={{Justice or Prejudice? Quantifying Biases in LLM-as-a-Judge}},
  author={Ye, Jiayi and Wang, Yanbo and Huang, Yue and Chen, Dongping and Zhang, Qihui and Moniz, Nuno and Gao, Tian and Geyer, Werner and Huang, Chao and Chen, Pin-Yu and Chawla, Nitesh V. and Zhang, Xiangliang},
  eprint={2410.02736},
  archivePrefix={arXiv},
  primaryClass={cs},
  year={2024},
  doi={10.48550/arXiv.2410.02736},
}

@inproceedings{chen2024humans,
  title={{Humans or LLMs as the Judge? A Study on Judgement Bias}},
  author={Chen, Guiming Hardy and Chen, Shunian and Liu, Ziche and Jiang, Feng and Wang, Benyou},
  booktitle={Proceedings of the 2024 Conference on Empirical Methods in Natural Language Processing},
  address={Miami, Florida, USA},
  publisher={Association for Computational Linguistics},
  year={2024},
  pages={8301--8327},
  doi={10.18653/v1/2024.emnlp-main.474}
}

@misc{wataoka2024self,
  title={{Self-Preference Bias in LLM-as-a-Judge}},
  author={Wataoka, Koki and Takahashi, Tsubasa and Ri, Ryokan},
  eprint = {2410.21819},
  archivePrefix={arXiv},
  primaryClass={cs},
  year={2024},
  doi={10.48550/arXiv.2410.21819}
}

@article{gu2026survey,
  title={{A Survey on LLM-as-a-Judge}},
  author={Gu, Jiawei and Jiang, Xuhui and Shi, Zhichao and Tan, Hexiang and Zhai, Xuehao and Xu, Chengjin and Li, Wei and Shen, Yinghan and Ma, Shengjie and Liu, Honghao and Wang, Saizhuo and Zhang, Kun and Lin, Zhouchi and Zhang, Bowen and Ni, Lionel and Gao, Wen and Wang, Yuanzhuo and Guo, Jian},
  journal={The Innovation},
  volume={7},
  number={6},
  pages={101253},
  year={2026},
  doi={10.1016/j.xinn.2025.101253}
}

@article{gao2025llm,
  title={{LLM-Based NLG Evaluation: Current Status and Challenges}},
  author={Gao, Mingqi and Hu, Xinyu and Yin, Xunjian and Ruan, Jie and Pu, Xiao and Wan, Xiaojun},
  journal={Computational Linguistics},
  volume={51},
  number={2},
  pages={661--687},
  year={2025},
  doi={10.1162/coli_a_00561}
}

@article{cheung2025large,
  title={{Large Language Models Show Amplified Cognitive Biases in Moral Decision-Making}},
  author={Cheung, Vanessa and Maier, Maximilian and Lieder, Falk},
  journal={Proceedings of the National Academy of Sciences},
  volume={122},
  number={25},
  pages={e2412015122},
  year={2025},
  doi={10.1073/pnas.2412015122}
}

@inproceedings{sumita2025cognitive,
  title={{Cognitive Biases in Large Language Models: A Survey and Mitigation Experiments}},
  author={Sumita, Yasuaki and Takeuchi, Koh and Kashima, Hisashi},
  booktitle={Proceedings of the 40th ACM/SIGAPP Symposium on Applied Computing},
  publisher={Association for Computing Machinery},
  address={New York, NY, USA},
  year={2025},
  pages={1009--1011},
  doi={10.1145/3672608.3707812}
}

@misc{huang2025anchoring,
  title={{An Empirical Study of the Anchoring Effect in LLMs: Existence, Mechanism, And Potential Mitigations}},
  author={Huang, Yiming and Bie, Biquan and Na, Zuqiu and Ruan, Weilin and Lei, Songxin and Yue, Yutao and He, Xinlei},
  eprint={2505.15392},
  archivePrefix={arXiv},
  primaryClass={cs},
  year={2025},
  doi={10.48550/arXiv.2505.15392}
}

@article{germani2025source,
  title={{Source Framing Triggers Systematic Bias in Large Language Models}},
  author={Germani, Federico and Spitale, Giovanni},
  journal={Science Advances},
  year={2025},
  volume={11},
  number={45},
  pages={eadz2924},
  doi={10.1126/sciadv.adz2924}
}

@misc{li2026preference,
  title={{Preference Leakage: A Contamination Problem in LLM-as-a-Judge}},
  author={Li, Dawei and Sun, Renliang and Huang, Yue and Zhong, Ming and Jiang, Bohan and Han, Jiawei and Zhang, Xiangliang and Wang, Wei and Liu, Huan},
  eprint={2502.01534},
  archivePrefix={arXiv},
  primaryClass={cs},
  year={2025},
  doi={10.48550/arXiv.2502.01534}
}

@misc{oriyad2025silent,
  title={{The Silent Judge: Unacknowledged Shortcut Bias in LLM-as-a-Judge}},
  author={Marioriyad, Arash and Rohban, Mohammad Hossein and Soleymani Baghshah, Mahdieh},
  eprint={2509.26072},
  archivePrefix={arXiv},
  primaryClass={cs},
  year={2025},
  doi={10.48550/arXiv.2509.26072}
}

@inproceedings{madaan2023selfrefine,
  title={{Self-Refine: Iterative Refinement with Self-Feedback}},
  author={Madaan, Aman and Tandon, Niket and Gupta, Prakhar and Hallinan, Skyler and Gao, Luyu and Wiegreffe, Sarah and Alon, Uri and Dziri, Nouha and Prabhumoye, Shrimai and Yang, Yiming and Gupta, Shashank and Majumder, Bodhisattwa Prasad and Hermann, Katherine and Welleck, Sean and Yazdanbakhsh, Amir and Clark, Peter},
  booktitle={Advances in Neural Information Processing Systems 36 (NeurIPS 2023)},
  publisher={Curran Associates Inc.},
  address={Red Hook, NY, USA},
  pages={46534--46594},
  year={2023},
  doi={10.52202/075280-2019}
}

@article{bao2025iterative,
  title={{Exploring Iterative Enhancement for Improving Learnersourced Multiple-Choice Question Explanations with Large Language Models}},
  author={Bao, Qiming and Leinonen, Juho and Peng, Alex Yuxuan and Zhong, Wanjun and Gendron, Ga\"{e}l and Pistotti, Timothy and Huang, Alice and Denny, Paul and Witbrock, Michael and Liu, Jiamou},
  journal={Proceedings of the AAAI Conference on Artificial Intelligence},
  volume={39},
  number={28},
  pages={28955--28963},
  publisher={Association for the Advancement of Artificial Intelligence},
  year={2025},
  doi={10.1609/aaai.v39i28.35164}
}

@misc{chhetri2025structsense,
  title={{StructSense: A Task-Agnostic Agentic Framework for Structured Information Extraction with Human-in-the-Loop Evaluation and Benchmarking}},
  author={Chhetri, Tek Raj and Chen, Yibei and Trivedi, Puja and Jarecka, Dorota and Haobsh, Saif and Ray, Patrick and Ng, Lydia and Ghosh, Satrajit S.},
  eprint={2507.03674},
  archivePrefix={arXiv},
  primaryClass={cs},
  year={2025},
  doi={10.48550/arXiv.2507.03674}
}

@inproceedings{yuksel2025multiagent,
  title={{A Multi-AI Agent System for Autonomous Optimization of Agentic AI Solutions Via Iterative Refinement And LLM-Driven Feedback Loops}},
  author={Yuksel, Kamer Ali and Castro Ferreira, Thiago and Al-Badrashiny, Mohamed and Sawaf, Hassan},
  booktitle={Proceedings of the 1st Workshop for Research on Agent Language Models (REALM 2025)},
  address={Vienna, Austria},
  publisher={Association for Computational Linguistics},
  year={2025},
  pages={52--62},
  doi={10.18653/v1/2025.realm-1.4}
}

@misc{han2026personalized,
  title={{Personalized Prediction of Perceived Message Effectiveness Using Large Language Model Based Digital Twins}},
  author={Han, Jasmin and Devkota, Janardan and Waring, Joseph and Luken, Amanda and Naughton, Felix and Vilardaga, Roger and Bricker, Jonathan and Latkin, Carl and Moran, Meghan and Chen, Yiqun and Thrul, Johannes},
  eprint={2602.19403},
  archivePrefix={arXiv},
  primaryClass={cs},
  year={2026},
  doi={10.48550/arXiv.2602.19403}
}

@misc{li2026grading,
  title={{Grading Scale Impact on LLM-as-a-Judge: Human-LLM Alignment is Highest on 0--5 Grading Scale}},
  author={Li, Weiyue and Zhao, Minda and Dong, Weixuan and Cai, Jiahui and Wei, Yuze and Pocress, Michael and Li, Yi and Yuan, Wanyan and Wang, Xiaoyue and Hou, Ruoyu and Lou, Kaiyuan and Zeng, Wenqi and Yang, Yutong and Du, Yilun and Wang, Mengyu},
  eprint={2601.03444},
  archivePrefix={arXiv},
  primaryClass={cs},
  year={2026},
  doi={10.48550/arXiv.2601.03444}
}

\end{document}